\documentclass{article}
\usepackage{spconf,amsmath,graphicx,hyperref}
\usepackage{amssymb}
\usepackage{iftex}
\ifxetex
  \usepackage{newtxtext}
\fi
\usepackage{booktabs,multirow,array}
\usepackage{pifont}
\usepackage{xcolor}
\usepackage{microtype}
\usepackage{url}
\usepackage{placeins}
\usepackage{balance}
\usepackage{stfloats}
\usepackage{float}
\hypersetup{hidelinks}

\makeatletter
\let\originalthebibliography\thebibliography
\renewcommand{\thebibliography}[1]{%
  \originalthebibliography{#1}%
  \setlength{\itemsep}{0pt}%
  \setlength{\parsep}{0pt}%
  \setlength{\topsep}{0pt}%
  \setlength{\baselineskip}{9pt}%
}
\makeatother

\newcommand{\cmark}{\ding{51}}
\newcommand{\xmark}{\ding{55}}

\newcommand{\norm}[1]{\left\lVert#1\right\rVert}

\title{WHEN RECOGNITION MEETS DISCOVERY:\\
LEARNING NEED NOT BE ENTANGLED IN ONE REPRESENTATION SPACE}

\makeatletter
\def\@maketitle{\newpage
 \null
 \vskip 2em
 \begin{center}
   {\large\bfseries \@title\par}
   \vskip .65em
   {\large\itshape \@name\par}
   \vskip .55em
   {\normalsize \@address\par}
 \end{center}
 \par
 \vskip .55em}
\makeatother

\name{Chenglin Liu$^{1}$, An Tian$^{1}$, Mingxuan Du$^{2}$, Shang Rui Yang$^{3}$}
\address{$^{1}$China University of Geosciences (Wuhan), Wuhan, China\\[1pt]
         $^{2}$The Hong Kong Polytechnic University, Hong Kong SAR, China\\[1pt]
         $^{3}$McGill University, Montreal, Canada}

\begin{document}
\ninept
\maketitle

\begin{abstract}
Generalized Category Discovery (GCD) aims to recognize known classes while
discovering novel categories from unlabeled data containing both old and novel
samples. Existing methods typically optimize recognition and discovery within a
shared representation space, implicitly assuming that the same representation
geometry can serve both objectives. Through controlled spectral analysis, we
uncover a striking asymmetry: old-class discrimination is increasingly
concentrated in the dominant singular directions, while novel-class structure is
better preserved in the complementary residual space. More importantly,
strengthening the former can even degrade the latter, revealing that recognition
and discovery compete for representation capacity rather than merely for
optimization weight. Motivated by this observation, we propose Dual-Space
Spectral Routing, which dynamically decomposes the representation via online SVD
and routes supervised recognition and unsupervised discovery to complementary
discriminative and general subspaces. We further exploit their distinct responses
to old and novel samples to construct a dual-space novelty signal, which guides
Subspace Relation Aggregation (SRA) for organizing candidate novel categories.
Extensive experiments across multiple GCD baselines, backbones, and benchmarks
demonstrate a consistently improved recognition--discovery balance without
introducing additional inference parameters. Code will be open source.
\end{abstract}

\begin{keywords}
Generalized category discovery, spectral routing, contrastive learning,
singular value decomposition
\end{keywords}

\section{Introduction}
\label{sec:intro}

Generalized Category Discovery (GCD)~\cite{vaze2022gcd} asks a model trained on
a labeled set of old classes to also partition an unlabeled set containing
both old and novel categories. It relaxes the closed-world assumption behind
standard recognition and extends novel category
discovery~\cite{han2019ncd,han2021autonovel,zhao2021dualranking,fini2021uno,li2023ncd},
where the unlabeled data are assumed to be entirely novel. Since old and novel
samples arrive mixed and unlabeled, the open-set portion cannot be separated in
advance~\cite{cao2022orca}.

Most recent methods build on a parametric classifier trained with
self-distillation~\cite{wen2023simgcd}. Improvements have come from mean-shift
refinement~\cite{cms}, Gaussian mixture
modeling~\cite{zhao2023gmm_gcd}, information
maximisation~\cite{chiaroni2023pim_gcd}, prompt
tuning~\cite{zhang2023promptcal,sptnet}, cross-instance
relations~\cite{pu2023dccl,cipr}, part-level~\cite{partco,selex,apl} and
scene-level~\cite{mos} priors, classifier debiasing~\cite{debgcd,flipclass},
hyperbolic geometry~\cite{hypcd}, and CLIP text embeddings~\cite{get}. The
choice of backbone and its fine-tuning schedule is also
significant~\cite{vaze2023norepresentation}. GCD has further been
studied under incremental~\cite{zhao2023incremental,kim2023proxyanchor},
active~\cite{ma2024activegcd}, federated~\cite{pu2024federatedgcd},
on-the-fly~\cite{du2023ocd,zheng2024phe}, and
cross-domain~\cite{wang2024hilo,rongali2024cdadnet,feng2025free} settings.

\begin{figure}[!t]
  \centering
  \includegraphics[width=0.8\linewidth]{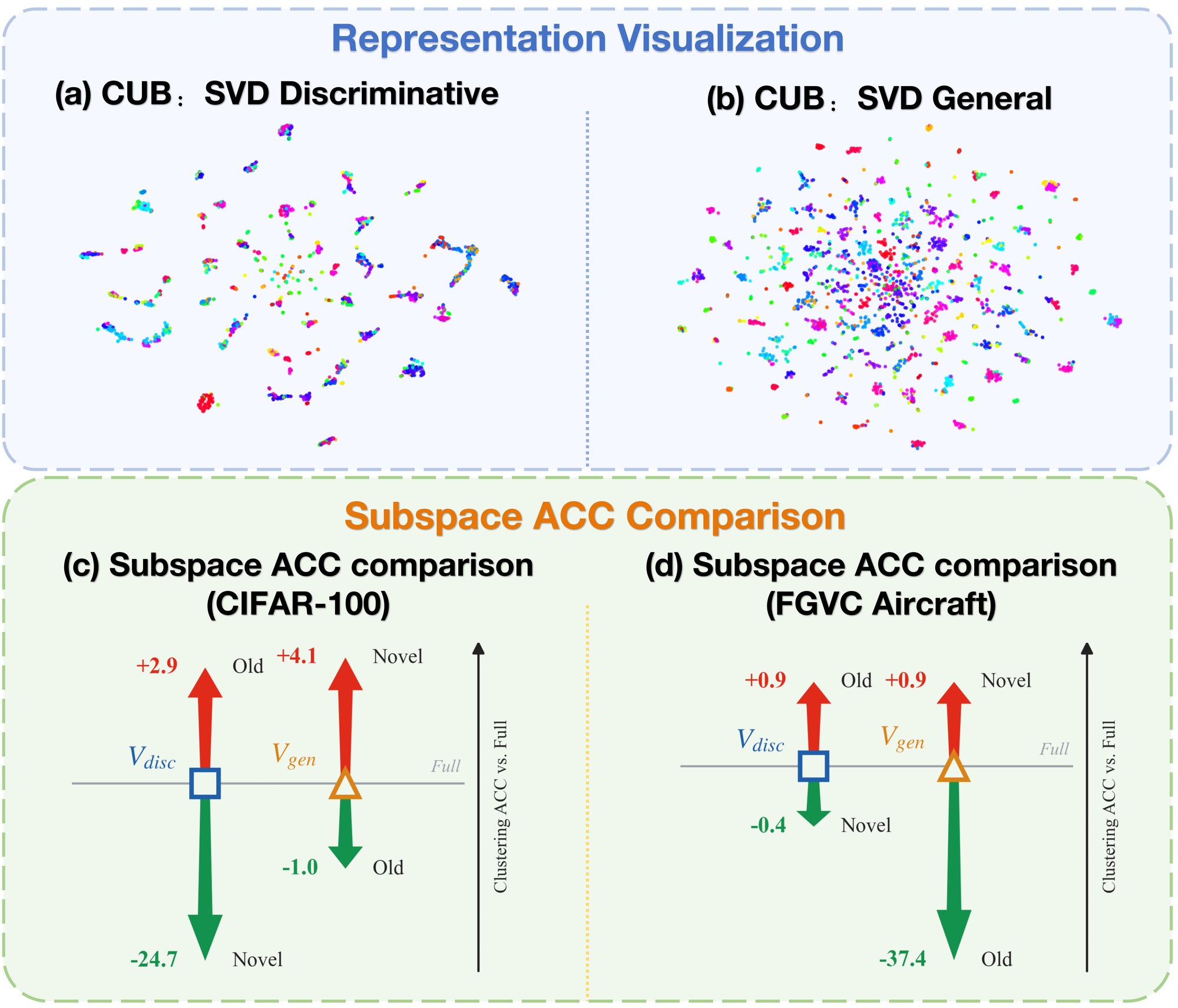}
  \vspace{-1mm}
  \caption{Recognition and discovery favor different subspaces.
  \textbf{Top:} t-SNE of CUB-200 features. (a) Old classes are compact in
  $V_{\rm disc}$; (b) far less separated in $V_{\rm gen}$. \textbf{Bottom:}
  accuracy change relative to the full representation (red: gain, green:
  loss). (c) $V_{\rm disc}$ costs $24.7$ points on novel classes; (d)
  $V_{\rm gen}$ costs $37.4$ on old ones.}
  \vspace{-5mm}
  \label{fig:motivation}
\end{figure}

\begin{figure*}[t]
  \centering
  \includegraphics[width=0.90\textwidth]{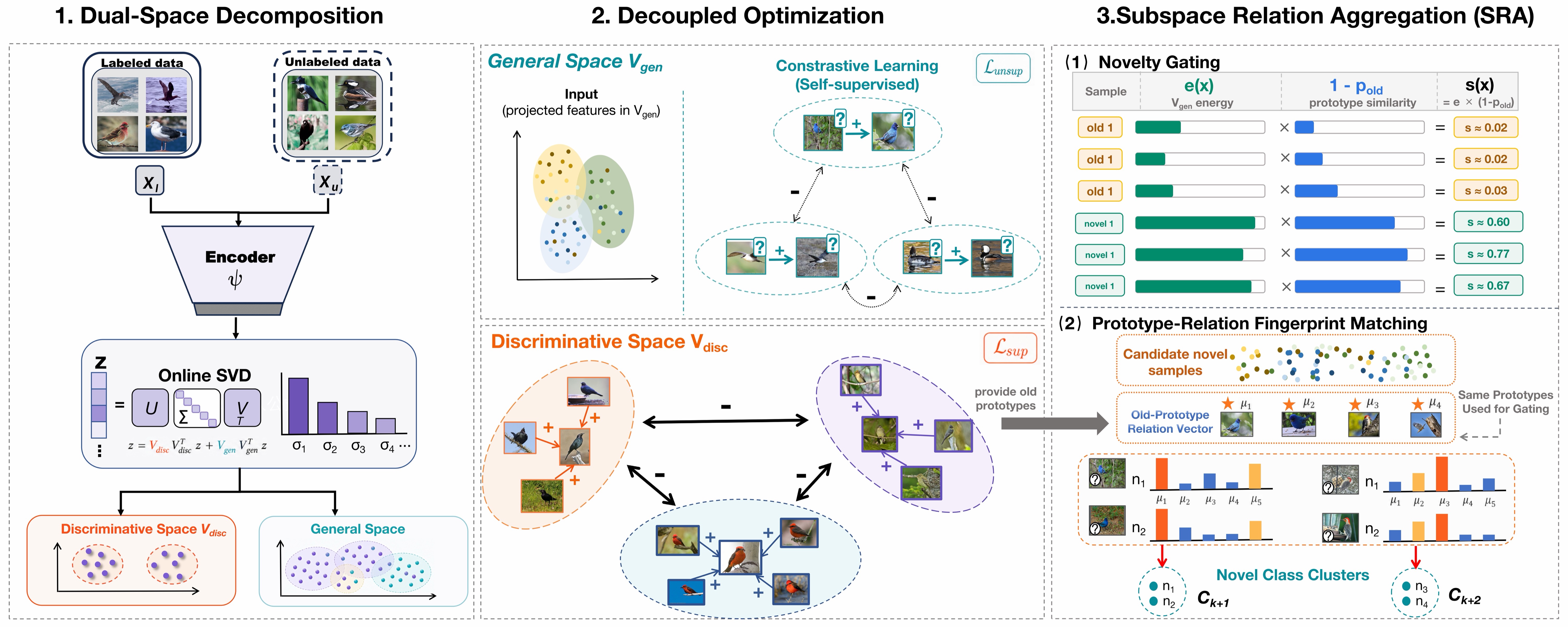}
  \vspace{-1mm}
  \caption{Overview of the proposed framework. \textbf{Left:} online SVD
  splits the representation into discriminative and general spaces.
  \textbf{Middle:} supervised and unsupervised objectives are routed to
  $V_{\rm disc}$ and $V_{\rm gen}$. \textbf{Right:} the novelty gate and SRA
  aggregate candidate novel samples via prototype-relation fingerprints.}
  \vspace{-5mm}
  \label{fig:framework}
\end{figure*}


\textcolor{black}{Most existing GCD methods nevertheless optimize supervised recognition and unsupervised discovery in a shared embedding space~\cite{wen2023simgcd,cms,zhao2023gmm_gcd,chiaroni2023pim_gcd,zhang2023promptcal,sptnet,pu2023dccl,cipr,debgcd,flipclass,mos,get,apl,hypcd,eagc}. Their mechanisms differ—from self-distillation~\cite{wen2023simgcd} and mean-shift refinement~\cite{cms} to debiasing~\cite{debgcd,flipclass}, part-level modeling~\cite{partco,selex,apl}, and alternative geometries~\cite{hypcd}—but they largely preserve the same underlying assumption: \textbf{\textit{one representation geometry can simultaneously become more discriminative for old classes and remain sufficiently expressive for novel ones.}} We question this assumption through a simple controlled experiment.}

\textcolor{black}{Specifically, we decompose a trained GCD representation~\cite{wen2023simgcd} along its singular spectrum and independently evaluate the leading and residual directions. The result is striking. As shown in Fig.~\ref{fig:motivation}, the leading singular directions strongly favor old-class recognition, whereas the complementary residual directions preserve disproportionately richer novel-class structure. On CIFAR-100, retaining the discriminative directions improves old-class accuracy by $2.9$ points but reduces novel-class accuracy by $24.7$ points; conversely, the residual space improves novel accuracy by $4.1$ points while only mildly affecting old classes. A similar asymmetry appears on FGVC Aircraft, where removing the dominant directions costs $37.4$ points on old classes while preserving or improving novel separation. \textbf{The directions that best explain the labeled taxonomy are therefore not necessarily the directions that best expose the unlabeled one.}}

\textcolor{black}{
This finding reveals a representation allocation conflict in GCD. Old-class supervision concentrates label-relevant information into dominant directions while suppressing variations that are less useful for the known taxonomy; yet these suppressed variations may still define novel categories. The recognition--discovery trade-off is therefore not merely about balancing supervised and unsupervised objectives, but also about where representation capacity is allocated. Conventional loss weighting changes how strongly the two objectives are optimized, while leaving them to compete within the same feature directions. Motivated by this observation, we propose \textbf{Dual-Space Spectral Routing}, which performs online SVD on the evolving representation and allocates its leading directions to a discriminative space for supervised recognition and its residual directions to a general space for unsupervised discovery. In this way, recognition can exploit label-aligned structure without forcing discovery to depend on the same representation geometry.}
\textcolor{black}{
The resulting decomposition further provides complementary evidence for novelty. Samples weakly matched to old-class prototypes in the discriminative space but retaining substantial energy in the general space are more likely to contain structure unexplained by the known taxonomy. We combine these cues into a soft novelty gate and introduce \textbf{Subspace Relation Aggregation (SRA)}, which organizes candidate novel samples according to their relations to old-class prototypes. Our method requires no auxiliary OOD classifier or additional inference parameters and can be directly incorporated into existing GCD frameworks. Experiments with SimGCD, CMS, and DebGCD under both DINO~\cite{caron2021dino} and DINOv2~\cite{oquab2024dinov2} demonstrate consistent improvements across generic and fine-grained benchmarks.
}

Our contributions are summarized as follows:
(i) We uncover a previously overlooked spectral asymmetry in GCD: old-class discrimination is concentrated in dominant representation directions, while residual directions preserve richer structure for novel-category discovery.
(ii) We propose Dual-Space Spectral Routing, together with a dual-space novelty signal and Subspace Relation Aggregation (SRA), to separate recognition and discovery across complementary spectral subspaces and organize candidate novel samples.
(iii) Extensive experiments across multiple GCD baselines and DINO/DINOv2 backbones demonstrate consistent improvements without additional inference parameters.

\section{Method}
\label{sec:method}

We route supervised recognition and unsupervised discovery into complementary
subspaces derived from a single online SVD of the current features: the leading
$k$ singular directions form the discriminative space and receive old-class
supervision, while the residual directions form the general space and receive
instance-level discovery. A dual-space novelty gate identifies candidate novel
samples, and Subspace Relation Aggregation (SRA) organizes them by their
prototype-relation fingerprints. The decomposition, prototypes, and gate are
refreshed periodically so the two spaces co-evolve with the representation. We
formalize the setting below, then detail the decomposition and its routing, and
finally the novelty gate and SRA.

\begin{table*}[!t]
\centering
\setlength{\tabcolsep}{0.20pt}
\renewcommand{\arraystretch}{0.84}
\caption{Results on six benchmarks with DINO ViT-B/16 in terms of ACC.}
\label{table:gcd}
\begin{tabular*}{\textwidth}{@{\extracolsep{\fill}}l *{18}{c}@{}}
\toprule
\multirow{2}{*}{Method} &
\multicolumn{3}{c}{CIFAR-100} & \multicolumn{3}{c}{ImageNet-100} &
\multicolumn{3}{c}{CUB-200} & \multicolumn{3}{c}{Stanford Cars} &
\multicolumn{3}{c}{FGVC Aircraft} & \multicolumn{3}{c}{Herbarium 19} \\
\cmidrule(lr){2-4}\cmidrule(lr){5-7}\cmidrule(lr){8-10}
\cmidrule(lr){11-13}\cmidrule(lr){14-16}\cmidrule(l){17-19}
& All & Old & New & All & Old & New & All & Old & New
& All & Old & New & All & Old & New & All & Old & New \\
\midrule
GCD~\cite{vaze2022gcd} & 73.0 & 76.2 & 66.5 & 74.1 & 89.8 & 66.3 & 51.3 & 56.6 & 48.7 & 39.0 & 57.6 & 29.9 & 45.0 & 41.1 & 46.9 & 35.4 & 51.0 & 27.0 \\
DCCL~\cite{pu2023dccl} & 75.3 & 76.8 & 70.2 & 80.5 & 90.5 & 76.2 & 63.5 & 60.8 & 64.9 & 43.1 & 55.7 & 36.2 & - & - & - & - & - & - \\
PromptCAL~\cite{zhang2023promptcal} & 81.2 & 84.2 & 75.3 & 83.1 & 92.7 & 78.3 & 62.9 & 64.4 & 62.1 & 50.2 & 70.1 & 40.6 & 52.2 & 52.2 & 52.3 & 37.0 & 52.0 & 28.9 \\
GPC~\cite{zhao2023gmm_gcd} & 77.9 & 85.0 & 63.0 & 76.9 & 94.3 & 71.0 & 55.4 & 58.2 & 53.1 & 42.8 & 59.2 & 32.8 & 46.3 & 42.5 & 47.9 & - & - & - \\
$\mu$GCD~\cite{vaze2023norepresentation} & - & - & - & - & - & - & 65.7 & 68.0 & 64.6 & 56.5 & 68.1 & 50.9 & 53.8 & 55.4 & 53.0 & - & - & - \\
SPTNet~\cite{sptnet} & 81.3 & 84.3 & 75.6 & 85.4 & 93.2 & 81.4 & 65.8 & 68.8 & \textbf{65.1} & 59.0 & 79.2 & 49.3 & 59.3 & 61.8 & 58.1 & - & - & - \\
\midrule
SimGCD~\cite{wen2023simgcd} & 80.1 & 81.2 & 77.8 & 83.0 & 93.1 & 77.9 & 60.3 & 65.6 & 57.7 & 53.8 & 71.9 & 45.0 & 54.2 & 59.1 & 51.8 & 44.0 & 58.0 & 36.4 \\
+ Ours & 81.3 & 82.2 & 79.6 & 83.5 & 93.5 & 80.2 & 66.1 & 76.6 & 60.9 & 60.4 & 80.6 & 50.6 & 56.0 & 62.3 & 52.8 & 44.1 & 58.0 & 36.7 \\
\addlinespace[1pt]
CMS~\cite{cms} & 82.3 & \textbf{85.7} & 75.5 & 84.7 & 95.6 & 79.2 & 68.2 & 76.5 & 64.0 & 56.9 & 76.1 & 47.6 & 56.0 & 63.4 & 52.3 & 36.4 & 54.9 & 26.4 \\
+ Ours & 82.1 & 83.6 & 79.1 & 84.9 & \textbf{95.8} & 79.4 & \textbf{68.8} & \textbf{76.7} & 64.8 & 57.3 & 76.3 & 48.2 & 56.2 & 63.2 & 52.8 & 38.2 & 56.0 & 27.3 \\
\addlinespace[1pt]
DebGCD~\cite{debgcd} & 83.0 & 84.6 & 79.9 & \textbf{85.9} & 94.3 & 81.6 & 66.3 & 71.8 & 63.5 & 65.3 & \textbf{81.6} & 57.4 & 61.7 & 63.9 & 60.6 & 45.8 & 58.9 & 38.5 \\
+ Ours & \textbf{83.7} & 85.2 & \textbf{80.3} & \textbf{85.9} & 94.1 & \textbf{81.7} & 67.4 & 72.8 & 63.8 & \textbf{65.5} & \textbf{81.6} & \textbf{57.8} & \textbf{62.0} & \textbf{64.3} & \textbf{60.7} & \textbf{46.2} & \textbf{59.2} & \textbf{39.0} \\
\midrule
{\color{green!55!black}\textbf{Avg. $\Delta$}}
& {\color{green!55!black}+0.57} & {\color{red!70!black}$-$0.17} & {\color{green!55!black}+1.93}
& {\color{green!55!black}+0.23} & {\color{green!55!black}+0.13} & {\color{green!55!black}+0.87}
& {\color{green!55!black}+2.50} & {\color{green!55!black}+4.07} & {\color{green!55!black}+1.43}
& {\color{green!55!black}+2.40} & {\color{green!55!black}+2.97} & {\color{green!55!black}+2.20}
& {\color{green!55!black}+0.77} & {\color{green!55!black}+1.13} & {\color{green!55!black}+0.53}
& {\color{green!55!black}+0.77} & {\color{green!55!black}+0.47} & {\color{green!55!black}+0.57} \\
\bottomrule
\end{tabular*}
\vspace{-5mm}
\end{table*}

\begin{table}[!t]
\centering
\small
\renewcommand{\arraystretch}{0.88}
\setlength{\tabcolsep}{0.20pt}
\caption{DINOv2 results on fine-grained benchmarks (ACC).}
\label{table:gcd_dinov2}
\vspace{-0.5mm}
\begin{tabular*}{0.98\columnwidth}{@{\extracolsep{\fill}}l*{9}{c}@{}}
\toprule
\multirow{2}{*}{Method} &
\multicolumn{3}{c}{CUB-200} & \multicolumn{3}{c}{Stanford Cars} &
\multicolumn{3}{c}{FGVC Aircraft} \\
\cmidrule(lr){2-4}\cmidrule(lr){5-7}\cmidrule(l){8-10}
& All & Old & New & All & Old & New & All & Old & New \\
\midrule
SPTNet~\cite{sptnet} & 76.3 & 79.5 & 74.6 & 72.3 & 82.0 & 67.5 & 68.0 & 75.2 & 60.9 \\
\midrule
SimGCD & 73.0 & 77.5 & 70.7 & 73.7 & 85.3 & 68.0 & 63.9 & 69.9 & 60.9 \\
+ Ours & 74.1 & 80.3 & 71.0 & 75.9 & 86.0 & 71.1 & 65.8 & \textbf{78.0} & 59.6 \\
\addlinespace[0pt]
CMS & 81.1 & \textbf{83.6} & 79.9 & 80.4 & \textbf{90.3} & 75.7 & 65.0 & 66.5 & 64.3 \\
+ Ours & \textbf{81.8} & 82.7 & \textbf{81.4} & \textbf{82.0} & 88.9 & \textbf{78.7} & 68.3 & 73.2 & 65.8 \\
\addlinespace[0pt]
DebGCD & 75.4 & 80.1 & 73.1 & 73.6 & 88.8 & 66.2 & 71.9 & 71.2 & 72.2 \\
+ Ours & 76.4 & 82.3 & 73.4 & 74.5 & 89.6 & 67.2 & \textbf{74.5} & 75.5 & \textbf{73.9} \\
\midrule
{\color{green!55!black}\textbf{Avg. $\Delta$}}
& {\color{green!55!black}+0.93} & {\color{green!55!black}+1.37} & {\color{green!55!black}+0.70}
& {\color{green!55!black}+1.57} & {\color{green!55!black}+0.03} & {\color{green!55!black}+2.37}
& {\color{green!55!black}+2.60} & {\color{green!55!black}+6.37} & {\color{green!55!black}+0.63} \\
\bottomrule
\end{tabular*}
\vspace{-4mm}
\end{table}

\subsection{Problem Setting}

Let $\mathcal D_L=\{(x_i,y_i)\}$ contain $C_L$ old classes, and let
$\mathcal D_U=\{x_i\}$ contain both old and novel samples. A DINO-pretrained
ViT-B/16~\cite{caron2021dino} and a projection head encode two augmentations of
each image as $z_i\in\mathbb R^d$ ($d=256$ in our DINO experiments).
Fig.~\ref{fig:framework} shows the resulting pipeline.

\subsection{Dual-Space Decomposition and Routing}

Let $Z\in\mathbb R^{N\times d}$ collect the current training features and let
$\bar Z=Z-\mathbf 1\mu^\top$ be its centered form, where $\mu$ is the feature
mean. One online SVD exposes the spectral structure of the representation:
\begin{equation}
 \bar Z=U\Sigma V^\top.
 \label{eq:svd}
\end{equation}
The right-singular vectors in $V$ come out ordered by how much feature variance
they explain. We cut that ordered basis at rank $k$ and read the two halves as
separate spaces:
\begin{equation}
 V_{\rm disc}=V_{:,1:k},\qquad V_{\rm gen}=V_{:,k+1:d}.
 \label{eq:split}
\end{equation}
Since both blocks are taken from a single orthonormal basis, they are exact
complements: whatever a feature loses in $V_{\rm disc}$ it keeps in
$V_{\rm gen}$. Each feature $z$ thus acquires two coordinate vectors,
\begin{equation}
 z^d=(z-\mu)V_{\rm disc},\qquad
 z^g=(z-\mu)V_{\rm gen},
 \label{eq:proj}
\end{equation}
where $z^d\in\mathbb R^{k}$ carries the dominant, largely old-class structure
and $z^g\in\mathbb R^{d-k}$ carries the residual variation that old-class labels
never constrained. This is exactly the division that
Fig.~\ref{fig:motivation} identifies as the site of the
recognition--discovery conflict, so we let each objective act where it belongs
and put supervision on $z^d$ and discovery on $z^g$:
\begin{equation}
 \mathcal L_{\rm base}=(1-\lambda)
 \mathcal L_{\rm NCE}(\operatorname{nrm}(z^g))
 +\lambda\mathcal L_{\rm SupCon}(\operatorname{nrm}(z_l^d),y_l),
 \label{eq:route}
\end{equation}
where $\operatorname{nrm}(\cdot)$ denotes $\ell_2$ normalization,
$\mathcal L_{\rm NCE}$ is the standard two-view InfoNCE
objective~\cite{oord2018cpc,chen2020simclr}, $\mathcal L_{\rm SupCon}$ is
supervised contrastive learning~\cite{khosla2020supcon}, $z_l^d$ is the labeled
subset of the discriminative coordinates, and $\lambda$ balances the two terms.
At the beginning of each update cycle, we re-extract training features and
refresh $\mu$, $V_{\rm disc}$, $V_{\rm gen}$, the old-class prototypes, and the
novelty threshold, so the two spaces co-evolve with the representation.

\subsection{Novelty-Gated Relation Aggregation}

Organizing novel categories requires knowing which unlabeled samples are novel.
Rather than train a separate OOD head, we read that signal directly off the
decomposition we already have. Two facts are informative. A novel sample is poorly explained by the
old-class directions, so an unusual share of its energy falls in the general
space; and it matches no old-class prototype particularly well.

We measure the first with an energy ratio. For old class $c$ we form a
normalized prototype in the discriminative space,
$p_c=\operatorname{nrm}(|\mathcal D_c|^{-1}\sum_{x_i\in\mathcal D_c}z_i^d)$, and
write $u=\operatorname{nrm}(z^d)$. The general-space share of a sample's energy
is then
\begin{equation}
 e(x)=\frac{\norm{z^g}_2^2}
 {\norm{z^d}_2^2+\norm{z^g}_2^2}.
 \label{eq:energy}
\end{equation}
Residual energy outside a principal subspace is a known out-of-distribution
cue~\cite{vim}; here it is read off the routed decomposition and used during
training rather than at test time.
The second is a soft assignment to the old classes,
\begin{equation}
 q_c(x)=\frac{\exp(u^\top p_c/\tau_p)}
 {\sum_{j=1}^{C_L}\exp(u^\top p_j/\tau_p)},
 \label{eq:oldprob}
\end{equation}
whose peak $p_{\rm old}(x)=\max_c q_c(x)$ says how confidently the sample can be
claimed by some known class. Multiplying the two gives a score that is large
only when both conditions hold at once:
\begin{equation}
 s(x)=e(x)\bigl(1-p_{\rm old}(x)\bigr).
 \label{eq:novelty}
\end{equation}
An absolute threshold on $s$ would be arbitrary, so we calibrate against the
labeled old classes, whose scores we can compute exactly. Taking the
$(1-\delta)$ quantile of their score distribution,
\begin{equation}
 \gamma=Q_{1-\delta}\!\left(\{s(x):x\in\mathcal D_L\}\right),
 \label{eq:thresh}
\end{equation}
places $\gamma$ just above where genuine old samples land. Comparing against it
through a sigmoid turns the score into a soft gate,
\begin{equation}
 \omega(x)=\sigma\!\left(\alpha(s(x)-\gamma)\right),
 \label{eq:gate}
\end{equation}
with $\alpha$ setting its sharpness. A soft gate lets each sample's contribution
track the evidence for it, so the grouping refines itself as the decomposition
sharpens.

Gated samples still have to be organized into categories. Our handle is that a
novel class, while matching no old class, tends to match the \emph{set} of old
prototypes in a characteristic way. We record that pattern as a fingerprint:
\begin{equation}
 \pi(x)=[u^\top p_1,\ldots,u^\top p_{C_L}].
 \label{eq:fingerprint}
\end{equation}
Samples of the same novel class should agree on this vector even though no
entry of it is large. Such relational signatures have also served pseudo-label
matching~\cite{rpc}; here they enter a pairwise loss whose weights come from
the gate above rather than from a separate classifier. To decide which pairs to compare, we first measure their
current proximity,
\begin{equation}
 \kappa_{ij}=\exp\!\left(\frac{\cos(u_i,u_j)}{\tau_u}\right),
 \label{eq:kappa}
\end{equation}
and then weight that proximity by how novel both samples look:
\begin{equation}
 W_{ij}=\omega(x_i)\,\omega(x_j)\,\kappa_{ij}.
 \label{eq:pairweight}
\end{equation}
A pair thus receives more weight when both samples appear novel and are
close. SRA then drives the weighted pairs toward a shared fingerprint:
\begin{equation}
 \mathcal L_{\rm SRA}=
 \frac{\sum_{i\ne j}W_{ij}\norm{\pi(x_i)-\pi(x_j)}_2^2}
 {\sum_{i\ne j}W_{ij}}.
 \label{eq:sra}
\end{equation}
Pairs receive stronger attraction when both appear novel and close, allowing
SRA to group candidates without assigning them to an old class. It is activated after a short warm-up, once the
decomposition has settled.

We retain each host's classification, self-distillation, and regularization
terms and augment its objective with the routed contrastive terms and SRA:
\begin{equation}
 \mathcal L=(1-\lambda)\mathcal L_{\rm NCE}
 +\lambda\mathcal L_{\rm SupCon}+\beta\mathcal L_{\rm SRA},
 \label{eq:total}
\end{equation}
where $\lambda$ balances recognition against discovery and $\beta$ weights SRA.
The SVD, the gate, and SRA are used only during training, so inference runs the
unmodified host encoder and adds no parameters.

\section{Experiments}
\label{sec:experiments}

\subsection{Setup}

\noindent\textbf{Datasets and protocol.}
We follow the standard GCD splits on CIFAR-100~\cite{krizhevsky2009cifar},
ImageNet-100~\cite{russakovsky2015imagenet}, CUB-200~\cite{wah2011cub},
Stanford Cars~\cite{krause2013cars}, FGVC Aircraft~\cite{maji2013fine}, and
Herbarium 19, and report Hungarian-matched clustering accuracy over All, Old,
and Novel classes~\cite{vaze2022gcd,wen2023simgcd}.

\noindent\textbf{Implementation.}
For DINO, we use ViT-B/16~\cite{dosovitskiy2021vit}, fine-tune its final
transformer block for 200 epochs with SGD (batch size 128, learning rate 0.1,
momentum 0.9), and set $\lambda=0.35$, $\tau_p=\tau_u=0.1$, and $\alpha=8$.
The basis, prototypes, and threshold are refreshed every five epochs, and SRA
starts after 20 epochs. We use $k=48$ on CUB-200, ImageNet-100 and Herbarium,
$k=34$ on Stanford Cars, $k=24$ on Aircraft, and $k=80$ on CIFAR-100;
$\beta=1.0$ on CUB, $0.5$ on CIFAR-100, and $0.1$ elsewhere; $\delta=0.05$
except $0.1$ on CIFAR-100. DINOv2 uses matched baseline/Ours pairs under
identical settings. All experiments are conducted on a single NVIDIA GeForce RTX 4090 GPU.

\subsection{Comparison with State-of-the-Art Methods}

\noindent\textbf{Results on DINO.}
Our method is plug-and-play and can be seamlessly integrated with the existing
objectives of SimGCD, CMS, and DebGCD, while retaining their host-specific
losses. Table~\ref{table:gcd} shows that Novel ACC improves on all six
benchmarks when averaged over the three hosts, indicating that the gain is not
tied to a particular GCD objective. The largest improvements appear on
fine-grained datasets: SimGCD + Ours improves All ACC by $5.8$ points on CUB-200
and $6.6$ points on Stanford Cars. These results support the benefit of
separating recognition and discovery into complementary representation spaces.

\noindent\textbf{Results on DINOv2.}
Table~\ref{table:gcd_dinov2} shows that stronger pretrained features do not
remove the benefit of spectral routing: the mean improvement over the three
hosts is positive for every dataset and metric. For CMS, All ACC increases by
$0.7$, $1.6$, and $3.3$ points on CUB-200, Stanford Cars, and FGVC Aircraft,
respectively, with consistent gains across backbones.

\subsection{Analysis and Ablation}

\begin{figure}[!tb]
  \centering
  \includegraphics[width=0.90\linewidth]{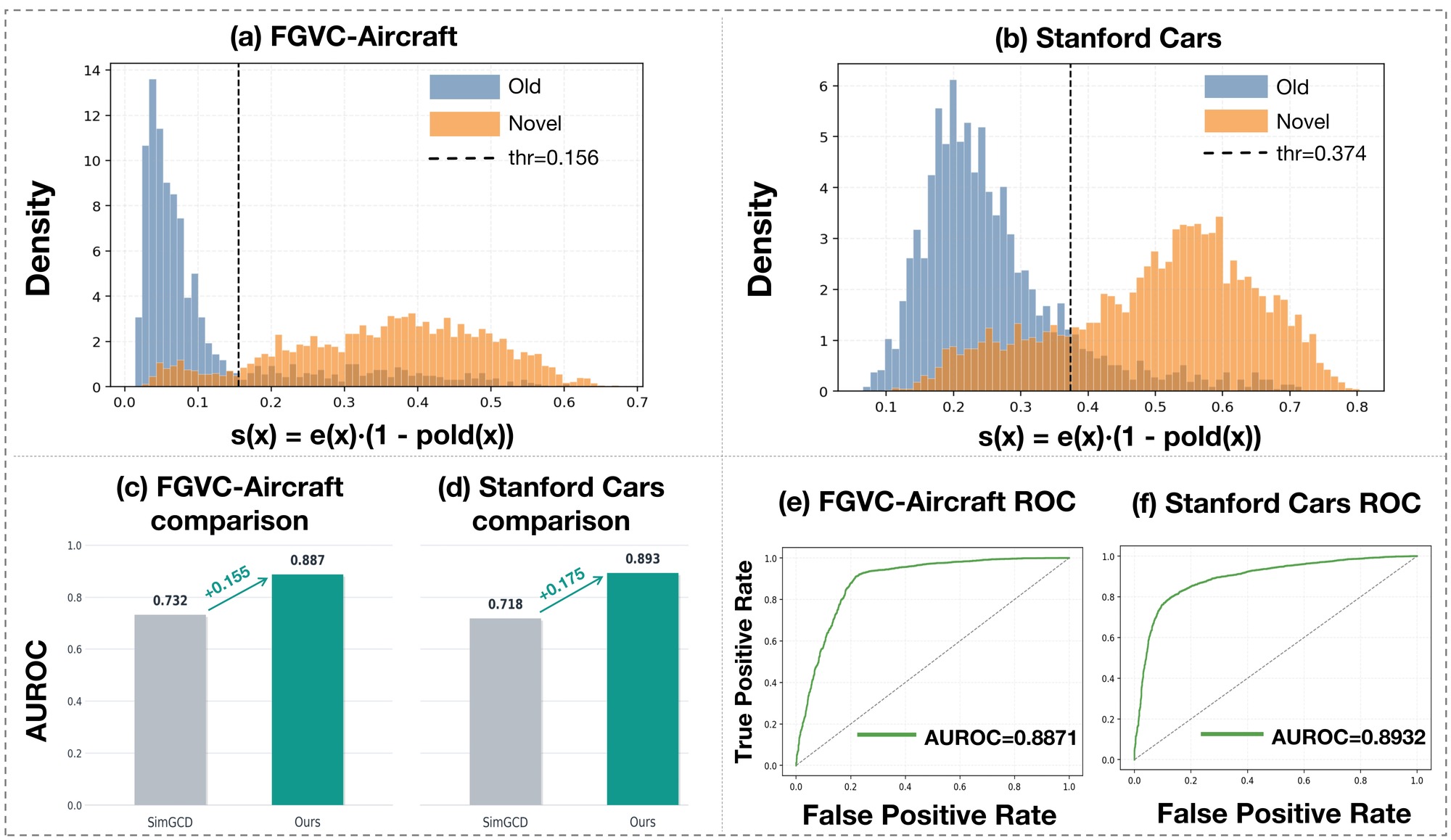}
  \vspace{-1mm}
  \caption{Old/novel separability of the novelty score on FGVC-Aircraft
  (left) and Stanford Cars (right). \textbf{Top:} score distributions.
  \textbf{Bottom:} AUROC against SimGCD and ROC curves.}
  \vspace{-5mm}
  \label{fig:auroc}
\end{figure}

\noindent\textbf{Old/Novel Separability.}
Figure~\ref{fig:auroc} shows that the proposed novelty score provides a clearer
old/novel boundary. AUROC increases from $0.732$ to $0.887$ on FGVC Aircraft
and from $0.718$ to $0.893$ on Stanford Cars.

\begin{table}[!t]
\centering
\caption{Ablation study on CUB-200.}
\label{tab:ablation_main}
\label{tab:supporting_ablation}
\vspace{-0.4mm}
\begingroup
\footnotesize
\renewcommand{\arraystretch}{0.84}
\setlength{\tabcolsep}{1.00pt}

\textbf{(a) Component ablation}
\vspace{0.1mm}

\begin{tabular*}{0.97\columnwidth}{@{\extracolsep{\fill}}cccccccccc@{}}
\toprule
\multicolumn{4}{c}{Components} & \multicolumn{3}{c}{SimGCD host} & \multicolumn{3}{c}{CMS host} \\
\cmidrule(lr){1-4}\cmidrule(lr){5-7}\cmidrule(l){8-10}
Dec. & Gate & SRA & Onl. & All & Old & New & All & Old & New \\
\midrule
\xmark & \xmark & \xmark & \xmark & 60.3 & 65.6 & 57.7 & 68.2 & 76.5 & 64.0 \\
\cmark & \xmark & \xmark & \xmark & 63.2 & 71.1 & 59.6 & 66.9 & 74.3 & 63.1 \\
\cmark & \cmark & \xmark & \xmark & 59.9 & 67.7 & 56.1 & 65.8 & 71.3 & 63.0 \\
\xmark & \cmark & \xmark & \xmark & 60.6 & 67.1 & 57.4 & 54.3 & 66.8 & 48.0 \\
\cmark & \cmark & \cmark & \xmark & 61.9 & 68.7 & 58.5 & 65.6 & 70.8 & 63.0 \\
\xmark & \cmark & \cmark & \xmark & 63.5 & 68.9 & 60.7 & 52.7 & 66.2 & 46.0 \\
\cmark & \cmark & \cmark & \cmark & \textbf{66.1} & \textbf{76.6} & \textbf{60.9} & \textbf{68.8} & \textbf{76.7} & \textbf{64.8} \\
\bottomrule
\end{tabular*}

\vspace{0mm}

\textbf{(b) Loss placement}
\vspace{0mm}

\begin{tabular*}{0.97\columnwidth}{@{\extracolsep{\fill}}lccccc@{}}
\toprule
Setting & $\mathcal{L}_{\rm sup}$ & $\mathcal{L}_{\rm unsup}$ & All & Old & New \\
\midrule
No routing & Full & Full & 60.3 & 65.6 & 57.7 \\
Both in Disc. & Disc. & Disc. & 63.6 & 74.7 & 58.0 \\
Both in Gen. & Gen. & Gen. & 65.1 & 74.5 & 60.4 \\
Gen. / Disc. & Gen. & Disc. & 65.5 & 70.2 & \textbf{63.2} \\
Ours & Disc. & Gen. & \textbf{66.1} & \textbf{76.6} & 60.9 \\
\bottomrule
\end{tabular*}
\vspace{-2.5mm}
\endgroup
\end{table}

\noindent\textbf{Component Ablation.}
Table~\ref{tab:ablation_main}(a) shows that routing alone raises SimGCD All ACC
from $60.3$ to $63.2$, while the complete online configuration achieves the
strongest overall performance. The full design reaches $66.1/76.6/60.9$
All/Old/New ACC on SimGCD and $68.8/76.7/64.8$ on CMS. This indicates that
routing, novelty gating, SRA, and online refresh work complementarily.

\noindent\textbf{Routing Evidence.}
Table~\ref{tab:ablation_main}(b) shows that the gain is not simply due to
projection into a lower-dimensional space. Routing supervised learning to the
discriminative space and unsupervised learning to the general space achieves
the best All and Old ACC, while the reversed assignment favors Novel ACC. This
supports the intended division of roles between the two complementary
subspaces.

\begin{figure}[!t]
  \centering
  \includegraphics[width=0.92\linewidth]{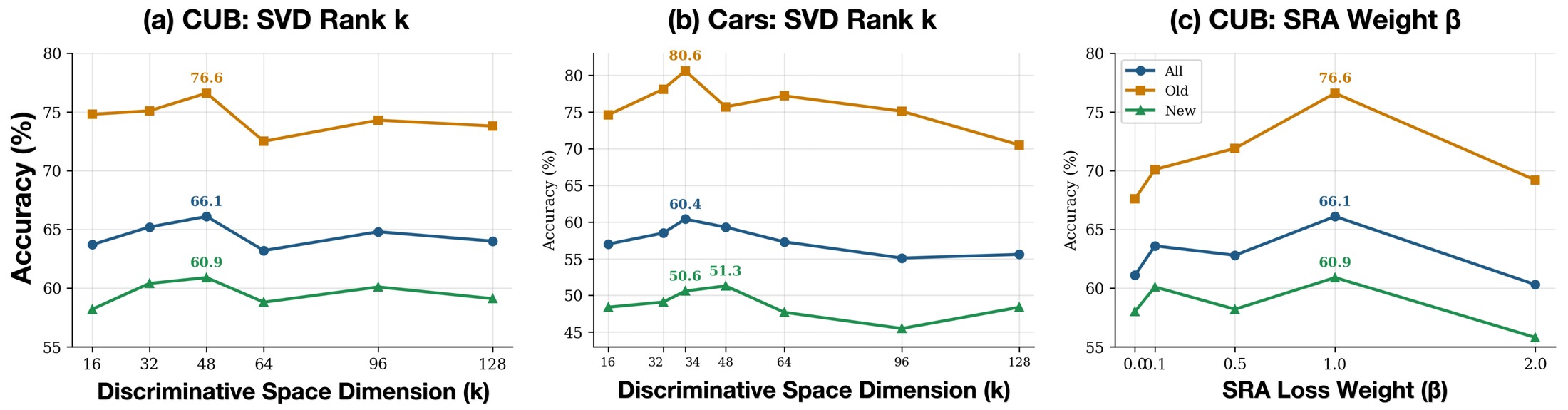}
  \vspace{-1mm}
  \caption{Hyperparameter analysis on CUB-200 and Stanford Cars.}
  \label{fig:hyperparam}
  \vspace{-5mm}
\end{figure}

\noindent\textbf{Hyperparameter Sensitivity.}
Figure~\ref{fig:hyperparam} shows that intermediate subspace ranks and moderate
SRA weights perform best. This is consistent with the goal of preserving
sufficient capacity for both recognition and discovery rather than allowing
either objective to dominate the representation.

\vspace{-2.5mm}
\section{Conclusion}

We identify a spectral asymmetry in GCD and exploit it with Dual-Space Spectral
Routing and SRA. Across three hosts, two backbones, and six benchmarks, the method
improves the recognition--discovery balance without additional inference parameters.

\clearpage
\bibliographystyle{IEEEbib}
\bibliography{references}

\end{document}